\documentclass[10pt,twocolumn,letterpaper]{article}

\usepackage{cvpr}              
\usepackage{booktabs}
\usepackage{multirow}
\usepackage{array}
\usepackage{listings}
\usepackage{xcolor}
\usepackage{graphicx}
\usepackage[most]{tcolorbox}
\usepackage{pgf-pie}
\usepackage{tikz}
\usetikzlibrary{arrows.meta}
\usetikzlibrary{shapes.geometric, positioning}
\usepackage{subcaption}
\usepackage{float}
\usepackage{adjustbox}
\usepackage{graphicx}
\usepackage{stfloats} 

\newtcolorbox{conversationbox}[2][]{%
    colback=white,
    colframe=black!70,
    boxrule=1pt,
    title=\textbf{#2},
    left=1mm, 
    right=1mm,
    top=1mm, 
    bottom=1mm,
    before skip=5pt,
    after skip=0pt,
    breakable=false,
    sharp corners=south, 
    #1
}

\newtcolorbox{subconversationbox}[1][]{%
    colback=white,
    colframe=black!70,
    boxrule=1pt,
    left=1mm, 
    right=1mm,
    top=1mm, 
    bottom=1mm,
    before skip=0pt, 
    after skip=5pt,
    breakable=false,
    sharp corners=north,
    #1
}

\definecolor{cvprblue}{rgb}{0.21,0.49,0.74}
\usepackage[pagebackref,breaklinks,colorlinks,allcolors=cvprblue]{hyperref}

\def\paperID{1} 
\def\confName{CVPR}
\def\confYear{2025}

\title{Behind Maya: Building a Multilingual Vision Language Model}

\author{
Nahid Alam$^{1*,2}$, 
\and 
Karthik Reddy Kanjula$^2$,
\and 
Surya Guthikonda$^{3,2}$,
\and 
Timothy Chung$^{4,2}$,
\and 
Bala Krishna S Vegesna$^5$,
\and 
Abhipsha Das$^2$,
\and 
Anthony Susevski$^2$,
\and 
Ryan Sze-Yin Chan$^6$,
\and 
S M Iftekhar Uddin$^2$,
\and 
Shayekh Bin Islam$^{7,2}$,
\and 
Roshan Santhosh$^8$,
\and 
Snegha A$^{9}$,
\and 
Drishti Sharma$^2$,
\and 
Chen Liu$^{10}$,
\and 
Isha Chaturvedi$^2$,
\and 
Genta Indra Winata$^{11*}$,
\and 
Ashvanth.S$^2$,
\and 
Snehanshu Mukherjee$^{12}$,
\and 
Alham Fikri Aji$^{13}$
\and 
\\
}

\begin{document}
\maketitle
\begin{abstract}
In recent times, we have seen a rapid development of large Vision-Language Models (VLMs). They have shown impressive results on academic benchmarks, primarily in widely spoken languages but lack performance on low-resource languages and varied cultural contexts. To address these limitations, we introduce Maya, an open-source Multilingual VLM. Our contributions are: 1) a multilingual image-text pretraining dataset in eight languages, based on the LLaVA pretraining dataset; and 2) a multilingual image-text model supporting these languages, enhancing cultural and linguistic comprehension in vision-language tasks. Code available at https://github.com/nahidalam/maya.
\end{abstract}    
\section{Introduction}
\label{sec:intro}

Recent progress in Large Language Models (LLMs) and vision encoders like CLIP \citep{radford2021learning} and SigLIP \citep{zhai2023sigmoid} has greatly advanced Vision-Language Models (VLMs). Models such as Flamingo \citep{alayrac2022flamingo}, LLaVA \citep{liu2023llava, liu2023improvedllava}, KOSMOS \citep{kosmos-g, peng2023kosmos}, Florence-2 \citep{xiao2024florence}, and Molmo \citep{deitke2024molmo} excel at image captioning, VQA, and reasoning. Qwen2-VL’s M-RoPE \citep{wang2024qwen2, su2021roformer} and PaLI’s joint modality and cross-lingual scaling \citep{chen2022pali, chen2023pali} further improve multimodal understanding. We acknowledge that although Aya Vision \citep{ayavision} showcases multilingual capabilities in 23 languages, it was released subsequent to the development of Maya.

In general though, VLMs still underperform in low-resource languages, struggling with cultural context and localized visual concepts \citep{joshi2020state}. A key reason is the lack of high-quality multilingual multimodal data. Most pretraining datasets—COCO \citep{lin2014microsoft}, Flickr30K \citep{young2014image}, LAION \citep{schuhmann2022laion}, Visual Genome \citep{krishna2017visual}, and LLaVA \citep{liu2023llava}—are English-centric, others \citep{elliott2016multi30k, thapliyal2022crossmodal} remain limited in scale and cultural coverage. To address these challenges, we introduce Maya, an open-source multilingual VLM that extends multimodal capabilities to eight languages.  Our key contributions:

\begin{enumerate}
	\item a novel multilingual image-text pretraining dataset consisting of \emph{550K} samples for future development of multilingual VLMs, 
	\item a new multilingual VLM that demonstrates improved performance in understanding cultural and linguistic nuances compared to PALO-7B \cite{maaz2024palo} on LLaVA-Bench-In-The-Wild \cite{liu2023llava}, offering a multilingual alternative to LLaVA \cite{liu2023llava}.
\end{enumerate}


\section{Dataset Creation and Filtering}
\label{sec:dataset}
\begin{figure*}[t]
  \centering
  \fbox{\includegraphics[width=0.95\textwidth]{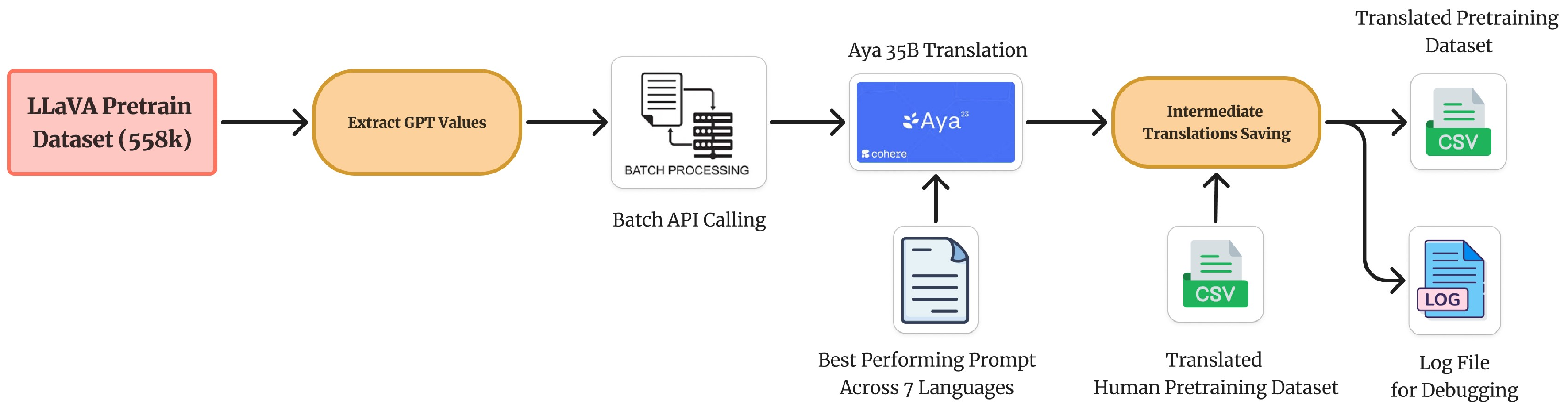}}
  \caption{Pretrain Dataset Preparation Process}
  \label{fig:createdata}
\end{figure*}
\subsection{Dataset Creation Methodology}

Recently, PALO \cite{maaz2024palo}  and Pangea \cite{yue2024pangea} have created a multilingual image-text dataset to build multilingual multimodal models. However, these multilingual datasets often suffer from data quality issues and distribution biases across languages. For example, in the PALO dataset, the distribution varies significantly between English and other languages \cite{elliott2016multi30k, hinck2024llava, maaz2024palo}. To address these limitations, we present a novel pre-training dataset tailored for LLaVA's architecture that is both multilingual and optimized for diverse language representation. Our dataset introduces rigorous processes for toxicity analysis, distribution balance, and quality control, ensuring consistent and reliable performance in multiple languages and modalities, as shown in Figure \ref{fig:createdata}. We expand the original English LLaVA dataset, which contains 550K samples, to include seven additional languages—Chinese, French, Spanish, Russian, Hindi, Japanese, and Arabic—yielding a total of 4.4 million samples, equally distributed across all eight languages. 

Our approach encompasses three components: 1) parallel dataset creation using a hybrid translation method, 2) prompt engineering optimization, and 3) scalable generation of pre-training datasets. This pipeline integrates multiple language models, such as \cite{gpt4o, team2023gemini, claude}, alongside specialized multilingual models like Aya 35B \cite{aryabumi2024aya}, to ensure quality, cross-lingual data suitable for multilingual applications.

\subsubsection{Multilingual LLaVA Pretrain Dataset}
Our Multilingual Pretrain dataset builds on the LLaVA dataset \citep{liu2023llava}, using image-text pairs and the corresponding GPT responses. Our approach started with sampling to select 30 diverse samples per language, guided by Length Analysis (LA), Flesch Reading Ease (FRE), and Flesch-Kincaid Grade Level (FKGL) metrics \citep{textstat} to obtain a representative GPT response in English. A cascaded translation and verification process ensures quality: initial translation using Google Translate, followed by back-translation and final human review with the help of \citep{claude, team2023gemini, achiam2023gpt} to generate the prompt engineering dataset in 8 languages.

\subsubsection{Prompt Engineering and Evaluation}

\begin{figure}[t]
  \centering
  \fbox{\includegraphics[width=0.9\linewidth]{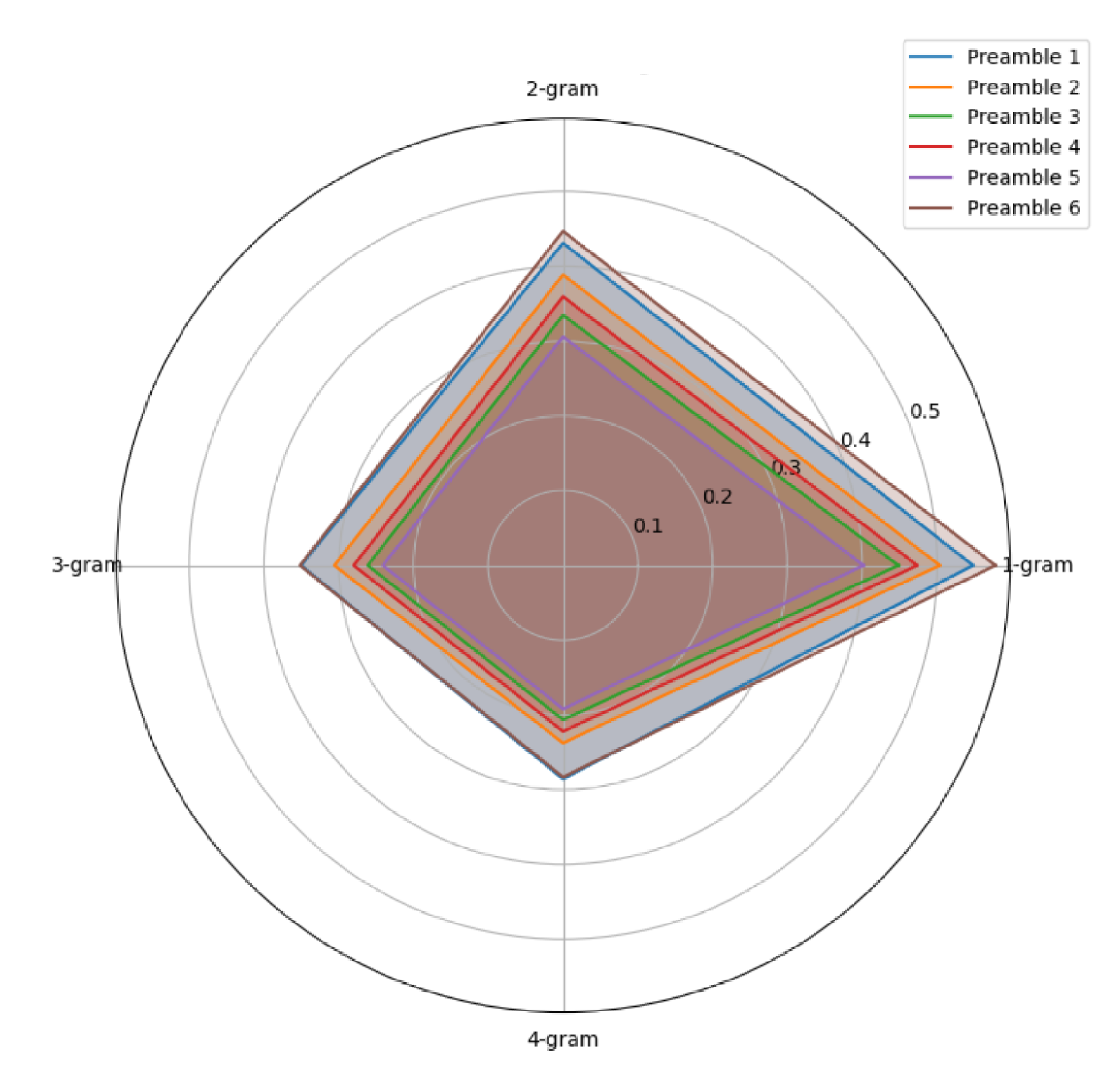}}
  \caption{Radar chart of N-gram averages by preamble}
  \label{fig:preamble_radar}
\end{figure}

During prompt engineering, we evaluate prompts per language using a BLEU score-based process. We construct a prompt evaluation dataset following Figure~\ref{fig:createdata}, translating six sample prompts into seven languages using Aya 35B. These translations are compared with reference translations of the prompt engineering dataset using BLEU \citep{papineni2002bleu} and N-gram scores \citep{shannon1948mathematical, brill-etal-1998-beyond}. In Arabic, Chinese, French, Hindi, Japanese, Russian and Spanish, Preamble 6 consistently produces the highest BLEU scores per Ngram (typically 0.4–0.5), showing a clear improvement from Type 5. Figure~\ref{fig:preamble_radar} shows Preamble 6 with the largest area in 1- to 4-grams, indicating better fidelity to the phrase level and consistent performance across languages. We adopt Preamble 6 as our final prompt template, shown in Listing~\ref{lst:translation_instructions}, and integrate it into our translation framework.
\lstset{
    basicstyle=\ttfamily\footnotesize,       
    backgroundcolor=\color{black!5},  
    frame=single,                     
    keywordstyle=\bfseries,           
    morekeywords={Input, Expected, Ensure, Note, Instructions, Examples},
    breaklines=true,                  
    breakatwhitespace=false,          
    columns=fullflexible              
}

\begin{figure}[t]
\begin{lstlisting}[caption={Translation Instructions}, label={lst:translation_instructions}]
## Instructions
You are an expert in translations. 
Your job is to translate the input to Japanese in
the given chat.

Ensure that:
{Specific Things to Consider while Translating}

Note: {Extra Constraints on Output Generation}

## Examples
### Example 1
Input:
{Input Sentence}
Expected Output:
{Translated Sentence}
\end{lstlisting}
\end{figure}


\subsubsection{Translation Framework Design}

Our translation framework uses the best preamble identified in the prompt engineering evaluation step. The prompt includes 1) standardized input-output formatting to maintain uniformity across languages, 2) example-driven instruction sets tailored for complex translations, and 3) integrated validation triggers to ensure quality control. Using Aya 35B translation capabilities, our framework achieves more than 0.47 average BLEU scores in seven languages. 

\subsubsection{Scalable Dataset Generation}

For large-scale dataset generation, we implement a batch processing pipeline integrated with Aya 35B API as shown in Figure \ref{fig:createdata}. The extracted GPT values from LLaVA Pretrain Dataset are passed through the Aya 35B batch-optimized API calling with intermediate translation checkpointing. The pipeline does necessary error handling and comprehensive logging for quality tracking. We implement version control for intermediate translations and maintain detailed debug logs, ensuring reproducibility and enabling systematic error analysis. The translation pipeline enables efficient processing of the 550K samples while maintaining translation quality.


\section{Multilingual Multimodal Instruction Tuning}
\label{sec:model}

\subsection{Model Architecture}

We draw inspiration from LLaVA 1.5 \cite{liu2023improvedllava} for model architecture. We employ the Aya-23 8B \cite{aryabumi2024aya} model as our LLM $f_{\phi}$ because of its multilingual capability. Aya-23 has 8 billion parameters, an 8K context window, and is trained across 23 languages. Our dataset, however, includes 8 of these 23 languages, aligning with our objective of optimizing Maya for a diverse yet focused linguistic range.

For the vision encoder, we opted for SigLIP\footnote{\texttt{siglip-base-patch16-256-multilingual} from \url{https://huggingface.co/google/siglip-base-patch16-256-multilingual}} \cite{zhai2023sigmoid} rather than CLIP \cite{radford2021learning}, which is traditionally used in LLaVA. This choice is motivated by SigLIP's strong performance, multilingual adaptability, and capacity for variable-length patch sizes. Unlike CLIP, SigLIP supports scalable positional embeddings, enabling it to accept inputs of varying dimensions through positional embedding interpolation. This flexibility makes SigLIP particularly suitable for our purpose. For each input image $X_v$, we get the visual features from SigLIP, $Z_v = g(X_v)$. A trainable projection matrix $W$ then converts the image features $Z_v$ into language features $H_v$.

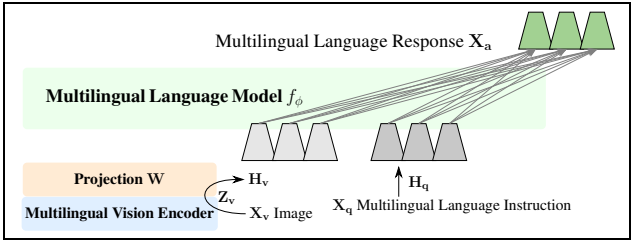
\begin{figure}[t]
    \centering
    \fbox{
    \resizebox{0.95\linewidth}{!}{ 
    \begin{tikzpicture}[
      node distance=1.4cm and 1.4cm,
      every node/.style={align=center},
      text height=2ex, text depth=.35ex,
      arrow style/.style={-{Stealth[length=3mm, width=2mm]}, thick, draw=black!50!white, opacity=0.5}
    ]
    
    \definecolor{lightgreen}{RGB}{236, 255, 236}
    \definecolor{lightblue}{RGB}{224, 239, 255}
    \definecolor{lightorange}{RGB}{255, 235, 212}
    \definecolor{greenbox}{RGB}{162, 209, 146}
    \definecolor{lightgrey1}{RGB}{230, 230, 230}
    \definecolor{lightgrey2}{RGB}{200, 200, 200}
    
    \node[fill=lightgreen, rounded corners, minimum width=17cm, minimum height=2cm, anchor=north west, align=left, text width=15.5cm] 
        (lm-bg) at (-6.2,0) {\LARGE \textbf{Multilingual Language Model }\( f_{\phi} \)};
    
    \node[fill=lightorange, rounded corners, minimum width=6.3cm, minimum height=1.1cm, anchor=south west] 
    (projection) at (-6.2,-4.2) {\Large\textbf{Projection} \( \mathbf{W} \)};
    \node[fill=lightblue, rounded corners, minimum width=6.3cm, minimum height=1.1cm, below=0cm of projection, align=left] 
    (vision-encoder) {\Large\textbf{Multilingual Vision Encoder}};
    
    \node[right=0.9cm of projection, anchor=west] (hv) {\Large \( \mathbf{H_v} \)};
    \node[right=0.9cm of vision-encoder, anchor=west] (xv) {\Large \( \mathbf{X_v} \) Image};
    
    \draw[-{Stealth[length=4mm, width=2.5mm]}, thick]
        (xv.west) 
        .. controls ++(-1.7,0.0) and ++(-1.7,0.0) 
        .. (hv.west);
    
    \node[above left= -0.1cm and 0.2cm of xv, align=center] (zv) {\Large \( \mathbf{Z_v} \)};
    
    \foreach \i in {0,1,2} {
        \node[fill=lightgrey1, draw=black, trapezium, trapezium angle=75, minimum width=1cm, minimum height=1.2cm, above=0.3cm of hv, xshift=\i cm] (hexagon\i) {};
    }
    
    \foreach \i in {3,4,5} {
        \node[fill=lightgrey2, draw=black, trapezium, trapezium angle=75, minimum width=1cm, minimum height=1.2cm, right=1.4cm of hexagon2, xshift=(\i-3)*1cm] (hexagon\i) {};
    }
    
    \node[below=0.4cm of hexagon4, anchor=north] (hq) {\Large \( \mathbf{H_q} \)};
    \node[below right=1.23cm and 1cm of hexagon3, anchor=north] (xq) {\quad\qquad\Large \( \mathbf{X_q} \) Multilingual Language Instruction};
    
    \draw[-{Stealth[length=4mm, width=2.5mm]}, thick] ([xshift=-1.2cm]xq.north) -- ++(0,1);
    
    \node[above left=2.6cm and 1.8cm of hexagon4, anchor=south] (xa) {\LARGE Multilingual Language Response \( \mathbf{X_a} \)};
    
    \foreach \i in {6,7,8} {
        \node[fill=greenbox, draw=black, trapezium, trapezium angle=75, minimum width=1cm, minimum height=1.2cm, right=0.9cm of xa, xshift=(\i-6)*1cm, yshift=0.4cm] (hexagon\i) {};
    }
    
    \foreach \j in {6,7,8} {
        \foreach \i in {0,1,2,3,4,5} {
            \draw[arrow style] (hexagon\i.north) -- (hexagon\j.south);
        }
    }
    
    \end{tikzpicture}}
    }
    \caption{Maya Architecture adapted from LLaVA \cite{liu2024llavanext}}
    \label{fig:onecol}
\end{figure}

\subsection{Pretraining}

For image-text alignment, we used a projection matrix $W$ that brings image features $X_v$ closer to language features. This projection matrix is a simple 2-layer MLP with GELU activation \citep{hendrycks2016gaussian}, as in LLaVA 1.5 \citep{liu2023improvedllava}. Although we experimented with 4- and 8-layer MLPs, the 2-layer configuration consistently achieved the lowest training loss. Advanced alignment techniques, such as gated soft-attention in Flamingo \cite{alayrac2022flamingo}, Q-Former in BLIP-2 \cite{li2023blip}, or pooling from MM1 \cite{mckinzie2024mm1} as alternatives to the projection layer, are set aside for future work. For each image $X_v$, we used the single-turn conversation data $$(X^1_q,X^1_a, \dots, X^T_q, X^T_a)$$ where $T$ is the total number of turns from LLaVA \cite{liu2023llava}. We pretrained Maya on the translated dataset in eight languages. Image inputs were cropped to 256x256 for compatibility with the SigLIP encoder. For training, we used 8xH100 GPUs with 80GB DRAM, per-device batch size of 32 and a global batch size of 256. A learning rate of 1e-3 and a cosine scheduler were applied during training. The pretraining process only trains the projection matrix and takes about 20 hours.

\subsection{Finetuning}

We instruction-finetuned our pretrained Maya model using the PALO 150K instruction-tuning dataset \citep{maaz2024palo}. During finetuning, we observed that Low Rank Adaptation (LoRA) \cite{hu2021lora} produced suboptimal results, particularly when both adapter matrices $A$ and $B$ were updated with the same learning rate \cite{hayou2024lora+}. Based on these findings, we opted against using LoRA and instead implemented full finetuning on 8xH100 GPUs. Our finetuning configuration used a per-device batch size of 16 and a global batch size of 128. Finetuning process took about 48 hours.

We kept both the vision encoder and the language encoders frozen during the training process. We did pretraining and finetuning for both versions of our dataset: the pretraining dataset translated into 7 languages.

\section{Results}
\label{sec:results}


\begin{table*}[!t]
    \centering
    \definecolor{highlight}{gray}{0.9}
    \definecolor{neg}{RGB}{0,0,255}
    \definecolor{pos}{RGB}{255,0,0}
    \begin{tabular}{lcccccccccc|c}
        \toprule
        Model & English & Chinese & French & Spanish & Russian & Japanese & Arabic & Hindi & Bengali & Urdu & Avg. \\
        \midrule
        \rowcolor{highlight}\textbf{Maya} (8B) & 61.5 & \underline{61.7} & 61.0 & 60.4 & \underline{62.2} & \underline{63.7} & \textbf{\underline{63.4}} & \underline{64.0} & 50.8 & 55 & \underline{60.4} \\
        LLaVA-7B & \underline{67.9} & 55.7 & \underline{62.4} & \underline{64.5} & 55.3 & 59.2 & 38.9 & 29.4 & 13.9 & 21.8 & 46.9 \\
        PALO-7B & 64.2 & 55.7 & 58.3 & 61.0 & 57.4 & 57.5 & 57.8 & 57.6 & \underline{51.7} & \underline{55.3} & 57.7 \\
        \midrule
        & \textcolor{neg}{-6.4} & \textcolor{pos}{+6.0} & \textcolor{neg}{-1.4} & \textcolor{neg}{-4.1} & \textcolor{pos}{+4.8} & \textcolor{pos}{+6.2} & \textcolor{pos}{+5.6} & 
        \textcolor{pos}{+6.4} & \textcolor{neg}{-0.9} & \textcolor{neg}{-0.3} & \textcolor{pos}{+2.7} \\
        \midrule
        LLaVA-13B & \textbf{\underline{69.5}} & \textbf{\underline{62.9}} & \textbf{\underline{67.5}} & 64.6 & 62.3 & \textbf{\underline{65.3}} & 37.2 & 27.8 & 20.4 & 22.1 & 49.9 \\
        PALO-13B & 65.5 & 62.1 & 66.4 & \textbf{\underline{65.9}} & \textbf{\underline{62.4}} & 60.6 & \underline{56.9} & \textbf{\underline{66.8}} & \textbf{\underline{53.5}} & \textbf{\underline{59.6}} & \textbf{\underline{61.9}} \\
        \midrule
        & \textcolor{neg}{-8.0} & \textcolor{neg}{-1.2} & \textcolor{neg}{-6.5} & \textcolor{neg}{-5.5} & \textcolor{neg}{-0.2} & \textcolor{neg}{-1.6} & \textcolor{pos}{+6.5} & \textcolor{neg}{-2.8} & 
        \textcolor{neg}{-2.7} & \textcolor{neg}{-4.6} & \textcolor{neg}{-1.5} \\
        \bottomrule
    \end{tabular}
    \caption{A comparison of LLaVA and PALO with Maya on eight languages adapted from LLaVA-Bench (In-the-Wild). Values \underline{underlined} indicate best performance within size class and values in \textbf{bold} indicate best performance across all models tested. We provide performance differences between Maya and the best competing model within the size classes where \textcolor{pos}{red} indicates where Maya is performing better and \textcolor{neg}{blue} indicates where Maya is performing worse than the best in the size class. ``Avg." represents the average over all the languages.}
    \label{tab:model-performance}
\end{table*}

\begin{table*}[!t]
\centering
\begin{tabular}{lcccccccccccccc}
\toprule
\textbf{GQA} & \textbf{VizWiz} & \textbf{ScienceQA} & \textbf{TextVQA} & \textbf{POPE-random} & \textbf{MMBench} & \textbf{MM-VeT} & \textbf{MME (P+C)} \\
\midrule
57.79\% & 34.92\% & 70.27\% & 47.01\% & 85.30\% & 71.10\% & 29.8 & 72.45\% \\
\bottomrule
\end{tabular}
\caption{Accuracy of Maya models on English Language across multiple benchmarks. Abbreviations: P+C = Perception + Cognition.}
\label{table:maya_eval_english}
\end{table*}

We evaluate Maya on the PALO multilingual benchmark \citep{maaz2024palo}, as shown in Table~\ref{tab:model-performance}. Although Maya was pretrained on only eight languages, it was finetuned on the PALO instruction dataset covering ten languages, and is thus evaluated on all ten. Maya outperforms all 7B models and is competitive with 13B models, surpassing LLaVA-13B on average and trailing PALO-13B by a small margin. Among the eight common languages, Maya outperforms PALO-7B in five. We attribute this to Maya's multilingual pretraining, in contrast to PALO’s English-only pretraining. Notably, Maya outperforms both PALO and LLaVA in Arabic for both size classes, likely due to its handling of root-based morphology and effective prompt design. We also evaluate Maya on English only dataset across various benchmarks as shown in Table ~\ref{table:maya_eval_english}.

\begin{figure}[t]
  \centering
  \fbox{\includegraphics[width=1\linewidth]{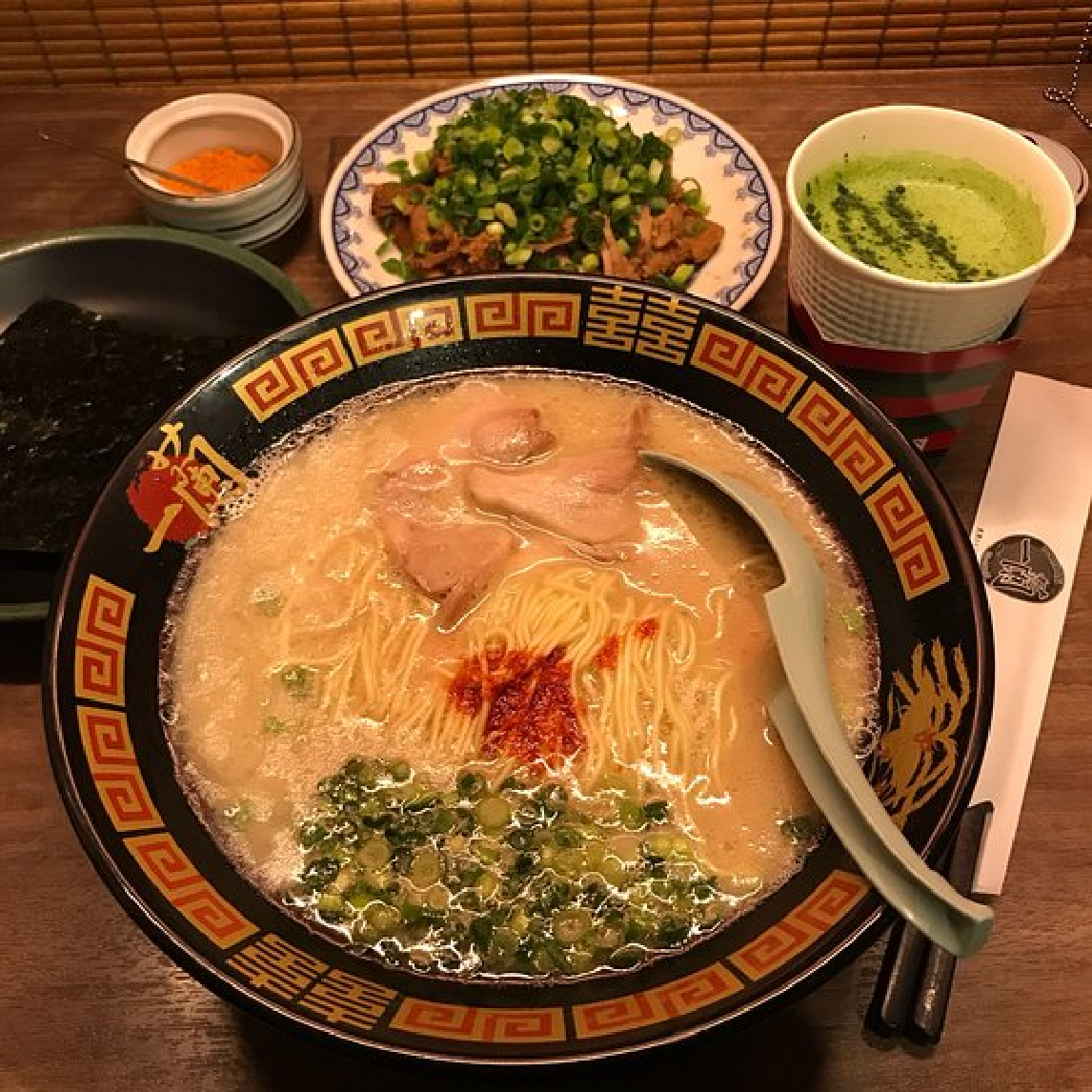}}
  \caption{Example image from LLaVA-Bench (In-the-Wild) \citep{liu2023llava}.}
  \label{fig:asianfood}
\end{figure}

\begin{figure}[t]
  \centering
  \fbox{\includegraphics[width=1\linewidth]{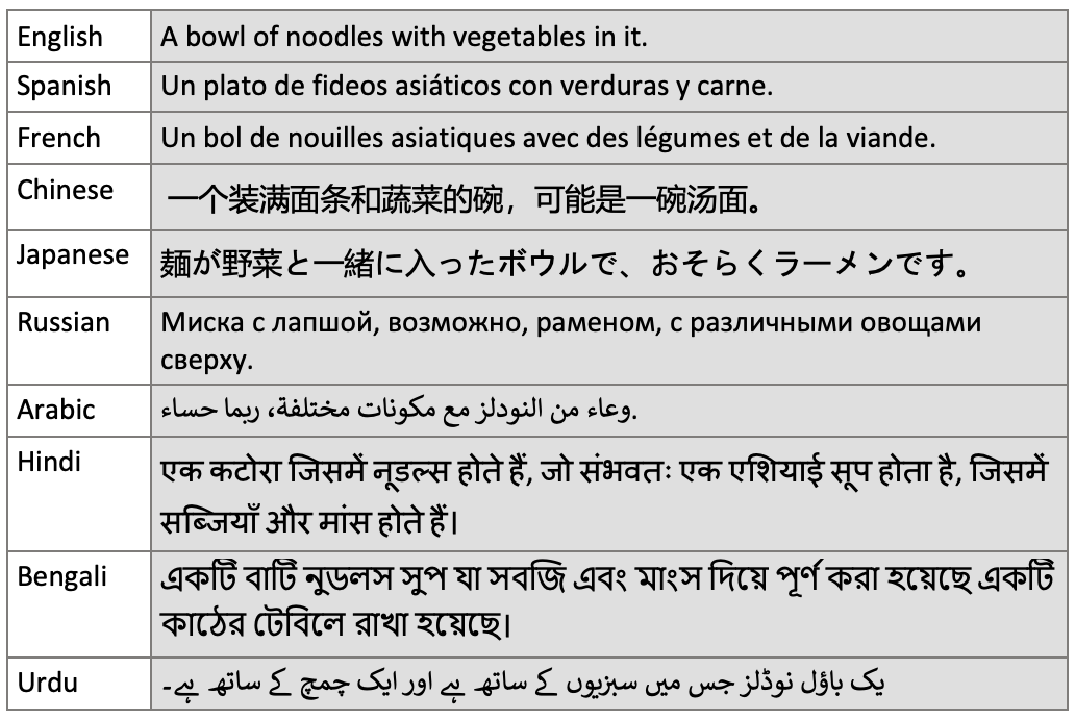}}
  \caption{Maya output for prompt (with image from Figure \ref{fig:asianfood}): Please describe the food in \{language\} in 1 sentence.}
  \label{fig:multilingual_output}
\end{figure}

Figure \ref{fig:multilingual_output} shows VQA outputs in various languages for Figure \ref{fig:asianfood}. The Bengali response is the most detailed, identifying both meat and the wooden table. Spanish, French, and Hindi mention the meat but miss the table. Chinese and Japanese outputs are similar to English in detail.

\section{Conclusion}
\label{sec:conclusion}

Maya enables high-quality multilingual, multimodal AI content generation, addressing gaps in low-resource languages. We curate data to minimize harmful content, though some traces may remain. Future work includes refining cross-modal alignment with alternative projection layers, unfreezing decoder layers for fine-tuning, and expanding pretraining to Bengali and Urdu. We also plan to scale the instruction-tuning dataset to 665K examples, improve translation via language-specific preambles, and benchmark on datasets like PangeaBench \citep{yue2024pangea} and CVQA \citep{romero2024cvqa} for broader, robust evaluation.

{
    \small
    \bibliographystyle{ieeenat_fullname}
    \bibliography{main}
}

\clearpage 
\section{Appendix}

\subsection{Author Affiliations}

\begin{itemize}
  \item[$^1$] Cisco Meraki
  \item[$^2$] Cohere Labs Community
  \item[$^3$] Indiana University Bloomington
  \item[$^4$] Imperial College London
  \item[$^5$] Georgia Institute of Technology
  \item[$^6$] The Alan Turing Institute
  \item[$^7$] Bangladesh University of Engineering and Technology
  \item[$^8$] University of Pennsylvania
  \item[$^9$] IIT Bombay
  \item[$^{10}$] TU Darmstadt
  \item[$^{11}$] Capital One
  \item[$^{12}$] IIT Dhanbad
  \item[$^{13}$] MBZUAI
\end{itemize}

\subsection{Acknowledgment}
We cannot express our gratitude enough to the Cohere Labs Community for bringing the open research community together which was instrumental to building Maya. The generous API credit for multilingual image-text translation from Cohere Labs Community has a direct impact on where the model exists today.  

\end{document}